\documentclass[letterpaper]{article} 
\usepackage{aaai24}  
\usepackage{times}  
\usepackage{helvet}  
\usepackage{courier}  
\usepackage[hyphens]{url}  
\usepackage{graphicx} 
\usepackage{natbib}  
\usepackage{caption} 
\usepackage{algorithm}
\usepackage{algorithmic}
\usepackage{color}
\usepackage{amsmath}
\usepackage{amssymb}
\usepackage{amsthm}
\usepackage{amsfonts}
\usepackage{multirow}
\usepackage{booktabs}
\usepackage[table]{xcolor}
\usepackage{newfloat}
\usepackage{listings}
\DeclareCaptionStyle{ruled}{labelfont=normalfont,labelsep=colon,strut=off} 
\floatstyle{ruled}
\newfloat{listing}{tb}{lst}{}
\floatname{listing}{Listing}
\title{Robust 3D Tracking with Quality-Aware Shape Completion}
\author {
    Jingwen Zhang\textsuperscript{\rm 1}\equalcontrib,
    Zikun Zhou\textsuperscript{\rm 2}\equalcontrib\thanks{Zikun Zhou and Wenjie Pei are corresponding authors.},
    Guangming Lu\textsuperscript{\rm 1},
    Jiandong Tian\textsuperscript{\rm 3},
    Wenjie Pei\textsuperscript{\rm 1}\footnotemark[2]\\
}
\affiliations {
    \textsuperscript{\rm 1}Harbin Institute of Technology, Shenzhen\\
    \textsuperscript{\rm 2}Peng Cheng Laboratory\\
    \textsuperscript{\rm 3}Shenyang Institute of Automation, Chinese Academy of Sciences\\
    \{jingwenz1022, zhouzikunhit\}@gmail.com, luguangm@hit.edu.cn,  tianjd@sia.cn, wenjiecoder@outlook.com
}

\usepackage{bibentry}

\begin{document}

\maketitle

\begin{abstract}
3D single object tracking remains a challenging problem due to the sparsity and incompleteness of the point clouds. Existing algorithms attempt to address the challenges in two strategies. The first strategy is to learn dense geometric features based on the captured sparse point cloud. Nevertheless, it is quite a formidable task since the learned dense geometric features are with high uncertainty for depicting the shape of the target object. The other strategy is to aggregate the sparse geometric features of multiple templates to enrich the shape information, which is a routine solution in 2D tracking. However, aggregating the coarse shape representations can hardly yield a precise shape representation. Different from 2D pixels, 3D points of different frames can be directly fused by coordinate transform, i.e., shape completion. Considering that, we propose to construct a synthetic target representation composed of dense and complete point clouds depicting the target shape precisely by shape completion for robust 3D tracking. Specifically, we design a voxelized 3D tracking framework with shape completion, in which we propose a quality-aware shape completion mechanism to alleviate the adverse effect of noisy historical predictions. It enables us to effectively construct and leverage the synthetic target representation. Besides, we also develop a voxelized relation modeling module and box refinement module to improve tracking performance. Favorable performance against state-of-the-art algorithms on three benchmarks demonstrates the effectiveness and generalization ability of our method.
\end{abstract}

\section{Introduction}
3D object tracking in LiDAR point clouds aims to predict the target position and orientation in subsequent frames, given the initial state of the target object. Existing 3D trackers~\cite{SC3D,P2B,STNet,PTT,PTTR} predominantly follow the Siamese tracking paradigm~\cite{SiamFC,SAOT}, which has achieved astonishing success in 2D tracking. The pioneering study SC3D~\cite{SC3D} calculates the feature similarities between the template and randomly sampled candidates to track the target. After that, many advanced techniques are introduced to improve 3D tracking performance, including end-to-end tracking framework~\cite{P2B}, transformer-based relation modeling~\cite{PTT, PTTR}, box-aware feature fusion~\cite{BAT}, and contextual information modeling~\cite{CXTrack,CMT}.

\begin{figure}[!t]
    \centering
    \includegraphics[width=1.0\columnwidth]{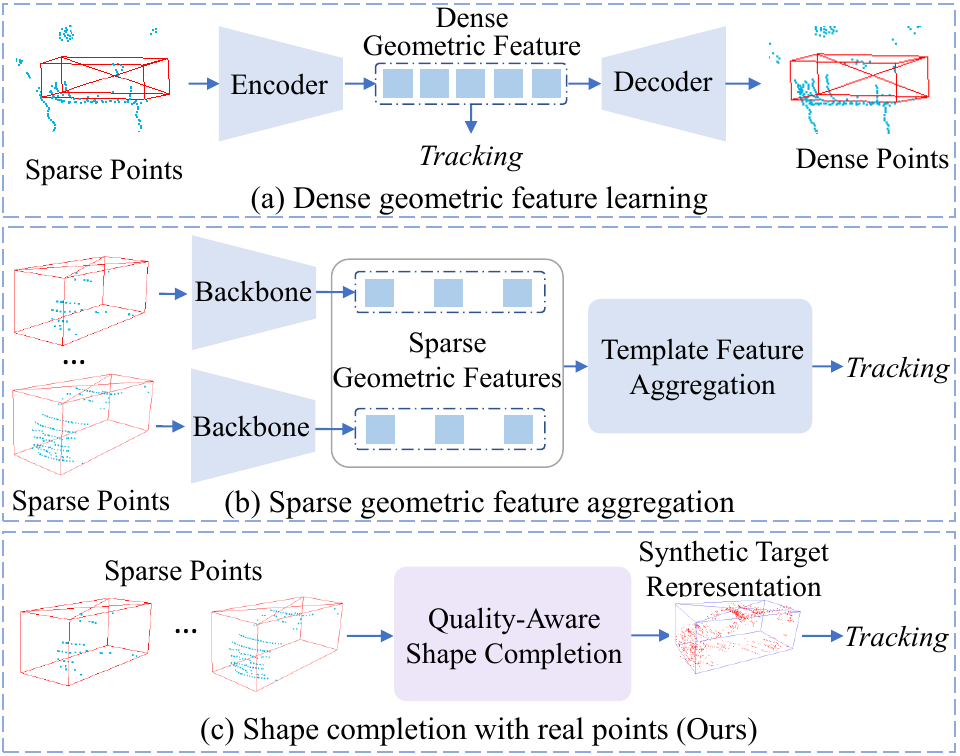}
    \caption{\textbf{Different methods for addressing the challenges of sparsity and incompleteness.} (a) Learning dense geometric features based on sparse points, which is a formidable task as the learned dense geometric features are with high uncertainty. (b) Aggregating the sparse geometric features of multiple templates, which is a sub-optimal solution as combining coarse shape representations can hardly obtain a precise shape. (c) Our method, which performs shape completion by adaptively fusing the real points of the target object from multiple frames to depict its shape precisely.}
    \label{Fig:introduction}
\end{figure}

Despite the great progress, many existing trackers~\cite{P2B,STNet,CXTrack,PTT,PTTR} pay less attention to the sparsity and incompleteness of the point clouds, which are usually caused by limited sensor capabilities and self-occlusion. For example, 51\% of cars in the KITTI~\cite{KITTI} dataset have less than 100 points. A typical challenging case is that only a few points of the template and the current target are overlapped due to the sparsity and incompleteness, in which accurately matching the template with the real target is quite difficult. As a result, these methods struggle to discriminate the target in extremely sparse and incomplete point clouds.
 
Several methods have been proposed to address the challenges of sparsity and incompleteness. SC3D~\cite{SC3D} and V2B~\cite{V2B} adopt a strategy of learning dense geometric features based on sparse point clouds, as shown in Figure~\ref{Fig:introduction} (a). However, such a learning task is quite formidable since the learned dense geometric features are with high uncertainty. The trackers take the risk of misleading by the inaccurate dense features. TAT~\cite{TAT} chooses to aggregate the sparse geometric features of multiple templates to obtain richer target shape information, which is a routine solution in 2D tracking~\cite{TMT, UpdateNet}, as shown in Figure~\ref{Fig:introduction} (b). Although this strategy allows the tracker to take more target points into account, aggregating the coarse shape representations extracted from sparse points can hardly generate a precise shape representation. Hence, this aggregation strategy is a sub-optimal solution for addressing the challenges of sparsity and incompleteness. 

Unlike the 2D image pixels, sparse 3D point clouds from different frames can be efficiently fused through coordinate transform to create a dense point cloud. Therefore, we propose to perform shape completion by fusing the target points from historical frames to construct a synthetic target representation for 3D tracking, as illustrated in Figure~\ref{Fig:introduction} (c). Herein the synthetic target representation consists of dense and complete point clouds depicting the shape of the target object precisely, enabling us to address the challenges of sparsity and incompleteness in 3D tracking.

In light of this idea, we design a robust 3D tracking framework that maintains a synthetic target representation by \textbf{s}hape \textbf{c}ompletion and performs 3D tracking in a \textbf{v}oxelized manner, termed SCVTrack. The tricky part of SCVTrack is that the synthetic target representation is sensitive to inaccurate historical predictions, and a noisy synthetic target representation can easily lead to tracking collapse. To alleviate the adverse effect of historical prediction errors, we propose a quality-aware shape completion module, which selectively fuses the well-aligned source points into the synthetic target representation. The shape completion naturally causes the imbalance between the point clouds of the template and the search area in terms of point density. It increases the difficulty of learning to model the relation between the two sets of point clouds. Therefore, we perform tracking based on the voxelized features instead of the point features to eliminate the imbalance. Besides, the voxelized tracking framework enables us to explicitly exploit the neighbor relation between voxels and is more computationally efficient. We also introduce a box refinement approach to further exploit the synthetic target representation to refine the target box, effectively improving tracking performance. 

To conclude, we make the following contributions: (1)~we propose a voxelized 3D tracking framework with shape completion to effectively leverage the real target points from historical frames to address the challenges of sparsity and incompleteness; (2)~we design a quality-aware shape completion mechanism, taking the quality of the points into account for shape completion to alleviate the adverse effect of historical prediction errors; (3)~we achieve favorable 3D tracking performance against state-of-the-art algorithms on three datasets, demonstrating the effectiveness of our method.

\section{Related Work}
\vspace{2mm}\noindent
\textbf{3D object tracking.}
Early 3D trackers~\cite{asvadi20163d, bibi20163d, liu2018context} based on RGB-Depth image pairs are vulnerable to lighting conditions that affect the RGB imaging quality. Recently, 3D tracking based on point clouds has drawn much more attention, as point clouds are robust to illumination changes. Most existing 3D trackers~\cite{SC3D,P2B,3D-SiamRPN,PTTR,STNet,CMT,MBPT} based on point clouds follow the Siamese tracking pipeline, which formulates 3D tracking as a template-candidate matching problem. Besides the Siamese tracking pipeline, M$^2$T~\cite{M2Track} recently proposes a motion-centric tracking paradigm, which directly predicts the target motion between two consecutive frames and achieves promising tracking performance. Despite the astonishing progress, the sparsity and incompleteness of 3D point clouds still plague these trackers.

An existing typical strategy to address the sparsity and incompleteness challenges is learning dense geometric features based on the given sparse point clouds. SC3D~\cite{SC3D} and V2B~\cite{V2B} are two methods following this strategy. However, such a learning task is quite challenging since the learned dense geometric features are with high uncertainty. As a result, these two methods only achieve limited tracking performance. A recently proposed approach, TAT~\cite{TAT}, adopts a multi-frame point feature aggregation strategy to enrich the shape information. Although it can alleviate the effect of sparsity and incompleteness, it is still a sub-optimal solution since aggregating the coarse shape representations extracted from sparse points can hardly generate a precise shape representation. Unlike the above methods, our method directly fuses the real target points from historical frames to construct a synthetic target representation for addressing the sparsity and incompleteness challenges.

\begin{figure*}[t]
\centering
    \includegraphics[width=1.0\textwidth]{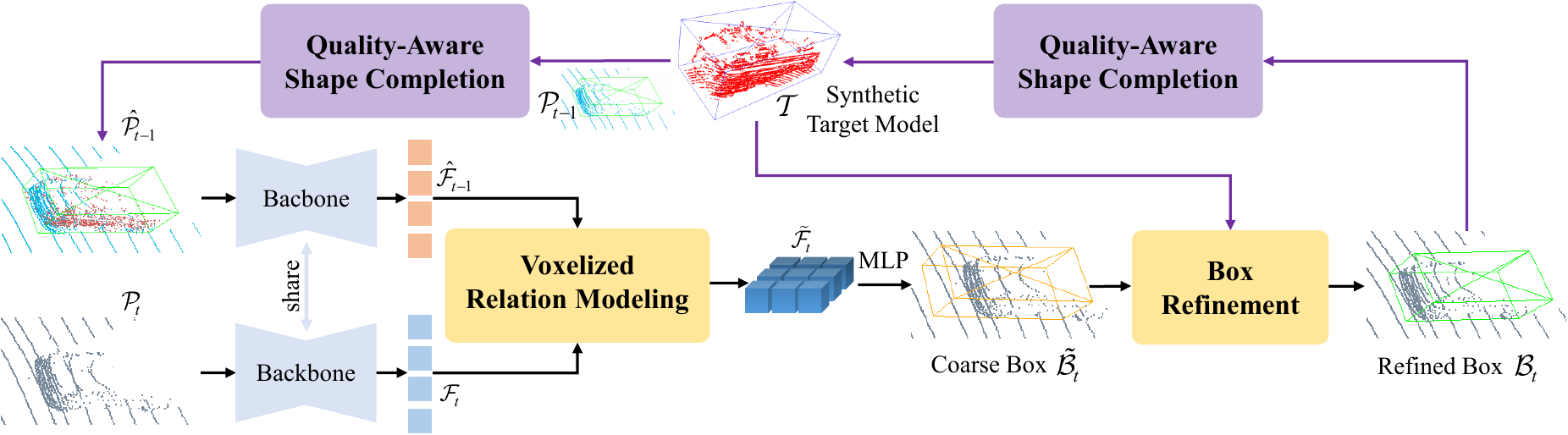}
    \caption{Overall framework of our SCVTrack, which mainly consists of a quality-aware shape completion module, a voxelized relation modeling module, and a box refinement module. It maintains a synthetic target representation $\mathcal T$ via quality-aware shape completion and performs 3D tracking in a voxelized manner. The red points in $\mathcal {\hat P}_{t-1}$ denote those coming from $\mathcal T$.}
\label{Fig:Framework}
\end{figure*}

\vspace{2mm}\noindent
\textbf{Voxel-based 3D vision.}
Most existing 3D trackers~\cite{SC3D,3D-SiamRPN,PTT} follow the point-based deep learning paradigm~\cite{li2018pointcnn, yang20203dssd}, performing tracking using the unordered point-based features. The voxel-based~\cite{liu2019point, qi2016volumetric, zhou2018voxelnet, yin2021center} learning paradigm is another popular way to process point data, which has been widely applied in 3D detection~\cite{zhou2018voxelnet, yan2018second, lang2019pointpillars, yin2021center}. However, it has rarely been explored in 3D tracking, except for V2B~\cite{V2B} and MTM~\cite{MTM}. This paradigm assigns the points into different voxel bins and extracts structured voxelized features from unordered point clouds. In this work, we resort to voxelized relation modeling to deal with the imbalance issue in terms of point density due to shape completion.

\section{Method}
\subsection{Problem Definition}
Given the initial 3D box of the target object, 3D tracking aims to estimate the target box in each subsequent frame. A 3D box is parameterized by its center position ($xyz$ coordinate), orientation (heading angle $\theta$ around the up-axis), and size (width $w$, length $l$, and height $h$). The size of the target object, even for non-rigid objects like pedestrians and cyclists, remains approximately unchanged in 3D tracking. Thus, we only predict the translation $(\Delta x, \Delta y, \Delta z)$ and the rotation angle $(\Delta \theta)$ of the target object between two consecutive frames, and then obtain the 3D box $\mathcal {B}_{t}$ at $t$-th frame by transforming $\mathcal {B}_{t-1}$ with the translation and rotation angle. 

\subsection{Overall Tracking Framework}
Figure~\ref{Fig:Framework} illustrates the overall framework of our SCVTrack. It mainly consists of the quality-aware shape completion, voxelized relation modeling, and box refinement modules. SCVTrack maintains a synthetic target representation $\mathcal{T}$ and performs tracking with it between two consecutive frames. Herein, the synthetic target representation is composed of dense and complete point clouds depicting the target shape precisely. We construct it by adaptively fusing the points belonging to the target from historical frames.

Suppose that the point clouds of the template and search area from two consecutive frames are denoted as $\mathcal{P}_{t-1}\in \mathbb{R}^{N_{t-1}\times 3}$ and $\mathcal{P}_t\in \mathbb{R}^{N_{t}\times 3}$, where $N_{t-1}$ and $N_{t}$ are the numbers of points. To localize the target in $\mathcal{P}_t$, our SCVTrack first completes $\mathcal{P}_{t-1}$ with the synthetic target representation $\mathcal {T}$ via quality-aware shape completion, yielding a completed template $\mathcal{\hat P}_{t-1}\in \mathbb{R}^{N'_{t-1}\times 3}$ with dense target points. Note that $N'_{t-1}$ is usually much larger than both $N_{t-1}$ and $N_{t}$ due to the shape completion. With $\mathcal{\hat P}_{t-1}$ and $\mathcal{P}_t$, SCVTrack adopts a shared backbone to extract their point features $\mathcal{\hat F}_{t-1}\in \mathbb{R}^{N'_{t-1}\times C}$ and $\mathcal{F}_t \in \mathbb{R}^{N_{t}\times C}$ without downsample, respectively, where $C$ is the feature dimension. Then SCVTrack voxelizes these point features and performs relation modeling between them to propagate the tracked target information from the template to the search area, generating the enhanced feature $\mathcal {\tilde F}_{t}$. An MLP regression head is constructed on top of $\mathcal {\tilde F}_{t}$ to predict a coarse box $\mathcal {\hat B}_{t}$. SCVTrack then performs box refinement with the guidance of $\mathcal {T}$ to obtain the refined box $\mathcal {B}_{t}$. After the tracking process, we use the new target points in $\mathcal {B}_{t}$ to update the synthetic target representation $\mathcal {T}$ via quality-aware shape completion. 

Note that we opt to complete $\mathcal {P}_{t-1}$ with $\mathcal T$ to obtain a dense template instead of directly using $T$ as the template. The rationale behind this design is that the target state in $\mathcal {P}_{t-1}$ is most similar to that in $\mathcal {P}_{t}$ in general, and completing $\mathcal {P}_{t-1}$ with $\mathcal T$ can not only obtain a dense template but also leverage all target points in $\mathcal {P}_{t-1}$ for tracking. 

\begin{figure*}[t]
\centering
    \includegraphics[width=0.95\textwidth]{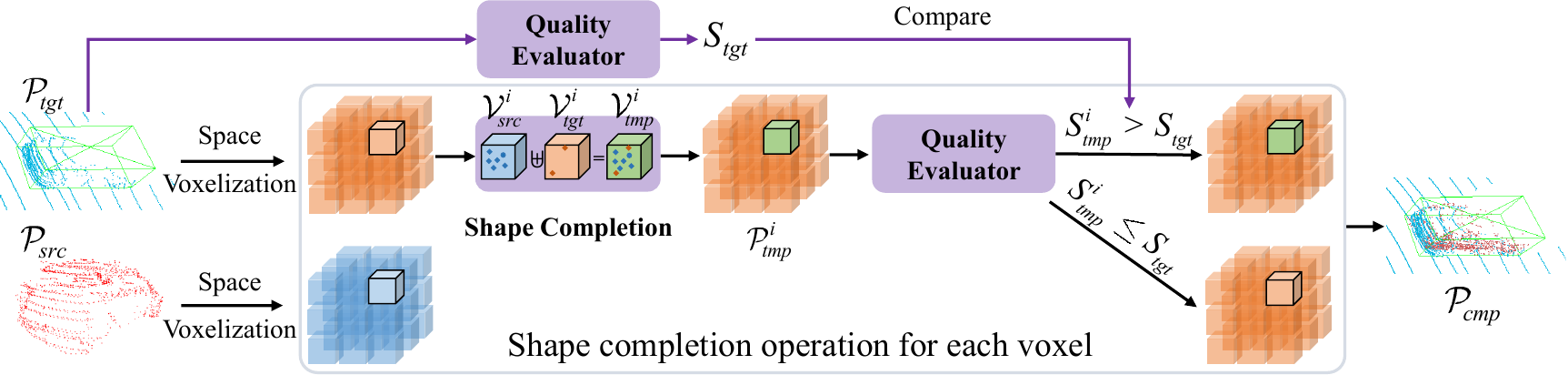}
    \caption{Illustration of the quality-aware shape completion module. $\uplus$ denotes the concatenation operation. This module performs selectively voxel-wise shape completion based on the output of the quality evaluator.}
\label{Fig:QASC}
\end{figure*}
\subsection{Quality-Aware Shape Completion}
\label{Sec:QASC}
The quality-aware shape completion module aims to adaptively fuse the source point cloud $\mathcal {P}_{src}$ with the point cloud $\mathcal {P}_{tgt}$ to be completed. For generating a dense template, $\mathcal{T}$ is treated as the source point cloud, and $\mathcal {P}_{t-1}$ is treated as the point cloud to be completed. In turn, for updating $\mathcal {T}$, $\mathcal{P}_{t}$ is treated as the source point cloud, and $\mathcal T$ is treated as the point cloud to be completed. Figure~\ref{Fig:QASC} shows the shape completion process taking the completion for generating a dense template as an example. In the above two completion process, the source point cloud $\mathcal {P}_{src}$ is always obtained based on predicted target states, which inevitably contains noisy points. Directly using all points in $\mathcal {P}_{src}$ for shape completion will lead to error accumulation and even tracking collapse. To address this issue, we propose to evaluate the quality of the point clouds and perform selectively voxel-wise shape completion conditioned on the quality score.

\vspace{2mm}\noindent
\textbf{Quality evaluator.}
To evaluate the quality of a point cloud, we design a quality evaluator consisting of a PointNet~\cite{PointNet} backbone and a three-layer MLP, which takes as input a point cloud and outputs a quality score. We formulate the quality evaluation task as a classification task. To be specific, we train the evaluator to differentiate the dense and well-aligned point clouds from the sparse and miss-aligned point clouds. To generate the required training samples, we first crop and center the points lying inside the object box from multiple frames. Then we concatenate these points to generate a dense and well-aligned point cloud as the positive sample. The negative sample is obtained by adding random position disturbance during concatenation or directly selecting a sparse point cloud from a certain frame. We use binary cross-entropy to train the quality evaluator. After training, the output logit is used as the quality score.

\vspace{2mm}\noindent
\textbf{Voxel-wise shape completion.}
With the quality evaluator, we design a voxel-wise shape completion strategy to selectively complete different parts of $\mathcal {P}_{tgt}$ to alleviate the adverse effect of the noisy points. To this end, we voxelize the 3D space and assign the points in $\mathcal {P}_{tgt}$ and $\mathcal {P}_{src}$ into the corresponding voxel bin, as shown in Figure~\ref{Fig:QASC}.
The points of $\mathcal {P}_{tgt}$ and $\mathcal {P}_{src}$ lying inside the $i$-th voxel bin are denoted by $\mathcal {V}^{i}_{tgt}$ and $\mathcal {V}^{i}_{src}$. Before shape completion, we first evaluate the quality of $\mathcal {P}_{tgt}$, obtaining its quality score $S_{tgt}$ as a reference. 
To complete the shape in the $i$-th voxel, we concatenate $\mathcal {V}^{i}_{src}$ with $\mathcal {V}^{i}_{tgt}$, yielding a dense point cloud $\mathcal {V}_{tmp}^{i}$ in the $i$-th voxel. We refer to the point cloud $\mathcal{P}_{tgt}$ whose $i$-th voxel is replaced with $\mathcal {V}_{tmp}^{i}$ as $\mathcal {P}_{tmp}^{i}$. Then we evaluate the quality of $\mathcal {P}_{tmp}^{i}$, obtaining a quality score $S_{tmp}^{i}$. After that, we compare $S_{tmp}^{i}$ with $S_{tgt}$ to judge whether the above completion in the $i$-th voxel improves the quality of the point cloud $\mathcal{P}_{tgt}$. Only when the quality is improved, we will update the points in the $i$-th voxel with $\mathcal {V}_{tmp}^{i}$. The above completion operation can be formulated as:
\begin{equation}
\begin{split}
    & \mathcal {V}_{tmp}^{i} = \mathcal {V}_{tgt}^{i} \uplus \mathcal{V}_{src}^{i};\\
    & S_{tmp}^{i} = \phi_{quality}(\mathcal{P}_{tmp}^{i});\\
    & \mathcal{V}_{cmp}^{i} = \left\{
    \begin{array}{rcl}
    \mathcal{V}_{tmp}^{i}, & & {if~S_{tmp}^{i} > S_{tgt};} \\
    \mathcal{V}_{tgt}^{i}, & & {else}.
    \end{array} \right.
\end{split}
\end{equation}
Herein, $\uplus$ denotes the concatenation operation, $\phi_{quality}$ refers to the quality evaluator, $\mathcal V_{cmp}^{i}$ denotes the points in the $i$-th voxel of the final completed point cloud $\mathcal {P}_{cmp}$. Note that the voxel-wise shape completion can be done in a single forward propagation, as the above completion operation for different voxels can be performed in parallel.

\begin{figure*}[t]
\centering
    \includegraphics[width=1.0\textwidth]{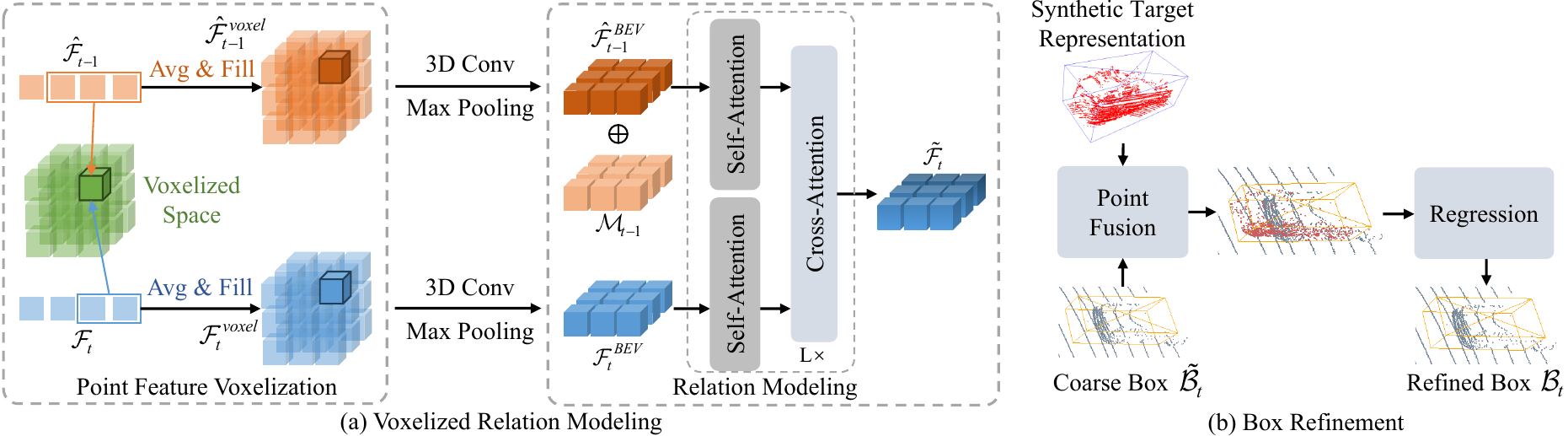}
    \caption{Illustration of the voxelized relation modeling and box refinement modules. $\oplus$ denotes element-wise summation.}
\label{Fig:Voxelized_Matching}
\end{figure*}

\subsection{Voxelized Relation Modeling}
Taking as input the point features $\mathcal {\hat F}_{t-1}$ and $\mathcal {F}_{t}$, relation modeling aims to propagate the target information from the previous frame to the current one, generating the enhanced feature $\mathcal {\tilde F}_{t}$ for localizing the target. 
As above-mentioned, the motivations that we opt for voxelized relation modeling lie in eliminating the imbalance between $\mathcal {\hat F}_{t-1}$ and $\mathcal {F}_{t}$ and exploiting the neighbor relation explicitly. To this end, we first voxelize 
the point feature and then perform relation modeling between the voxelized feature, as shown in Figure~\ref{Fig:Voxelized_Matching}.

\vspace{2mm}\noindent
\textbf{Point feature voxelization.} 
We convert the point features $\mathcal {\hat F}_{t-1}$ and $\mathcal {F}_{t}$ into the voxelized representations $\mathcal {\hat F}^{vxl}_{t-1}$ and $\mathcal {F}_{t}^{vxl}$, respectively, by averaging the features of the points lying inside the same voxel bin. Then we apply shared 3D convolution layers to aggregate the shape information in the adjacent feature voxels to enhance the voxelized feature representations. Similar to~\cite{V2B}, we perform max-pooling on these voxelized features along the $z$-axis to obtain the dense bird's eye view (BEV) features $\mathcal {\hat F}^{bev}_{t-1}\in \mathbb{R}^{H\times W \times C}$ and $\mathcal {F}^{bev}_{t}\in \mathbb{R}^{H\times W \times C}$ to alleviate the adverse effect the empty voxels.

\vspace{2mm}\noindent
\textbf{Relation modeling.}
Considering that $\mathcal {P}_{t-1}$ contains both the target and background points, we introduce a learnable target mask $\mathcal M_{t-1}\in \mathbb{R}^{H\times W \times C}$ to embed the target state information into $\mathcal {\hat F}_{t-1}^{bev}$ before relation modeling. Technically, we introduce three learnable vectors indicating the three different positional states of a voxel, which are lying inside the box, outside the box, and across the box boundary. Then we generate the mask $\mathcal M_{t-1}$ according to the 2D projection box (along the $z$-axis) of the 3D box $\mathcal B_{t-1}$.

Inspired by recent advances~\cite{OSTrack, JNLT} in 2D tracking, we adopt an attention-based method to propagate the target information from $\mathcal {\hat F}_{t-1}^{bev}$ to $\mathcal {F}_{t}^{bev}$. As shown in Figure~\ref{Fig:Framework}, a shared self-attention layer is first employed to model the intra-frame voxel relation. Then a cross-attention is used to model the cross-frame voxel relation, where the feature of the current frame is used as query and the feature of the previous frame is used as key and value. This process can be formulated as:
\begin{equation}
\mathcal {\tilde F}_t=\psi_{ca}(\psi_{sa}(\mathcal {\hat F}_{t-1}^{bev} \oplus \mathcal {M}_{t-1} \oplus  \mathcal {E}), \psi_{sa}(\mathcal {F}_{t}^{bev} \oplus  \mathcal {E})),
\label{Eq:matching}
\end{equation}
where $\psi_{sa}$ and $\psi_{ca}$ denote the self-attention and cross-attention, respectively. $\oplus$ means element-wise summation, and $\mathcal E$ refers to the position embedding. Note that we omit the flatten and reshape operation in Eq.~\ref{Eq:matching}, and this attention architecture is repeated $L$ times.

\subsection{Box Refinement}
The box refinement module aims to refine the coarse box $\mathcal {\tilde B}_{t}$ with the guidance of the dense geometric information in $\mathcal {T}$. To this end, we first fuse the dense points in $\mathcal {T}$ into the coarse box $\mathcal {\tilde B}_{t}$ by coordinate transform, obtaining a new point could $\mathcal {\hat P}_{t}$ depicting the target object with dense points. Note that $\mathcal T$ is nearly aligned with the target in the search area after the coordinate transform. The offset between the coarse box $\mathcal {\tilde B}_{t}$ and the real target object will affect the smoothness of $\mathcal {\hat P}_{t}$. Based on this principle, we deploy a PointNet backbone following an MLP on top of $\mathcal {\hat P}_{t}$ to regress the above-mentioned offset to refine the target box. Such a process is akin to performing registration between the synthetic target representation $\mathcal T$ and the target in the search area. But our method is different from RDT~\cite{RDT}, which uses registration to provide spatially aligned points for Siamese matching. Our method exploits registration to refine the coarse box which is predicted via voxelized matching.

\subsection{End-to-end Modeling Learning}
Our framework consists of two learnable parts: the quality evaluator and the remaining tracking model, which are trained separately. The quality evaluator is end-to-end trained as above-mentioned. The tracking model is end-to-end trained with pairs of consecutive frames. We impose smooth-$l1$ loss~\cite{girshick2015fast} on both the coarse and refined boxes to supervise the learning of the tracking model. Note that the synthetic target representation used in tracking model learning is pre-calculated with grounding truth.

\section{Experiments}
\subsection{Experimental Setup}
\noindent
\textbf{Implementation details.}
We use a modified PointNet++~\cite{qi2017pointnet++} as our backbone, which is tailored to contain three set-abstraction (SA) layers and three feature propagation (FP) layers. In the three SA layers, the sample radiuses are set to 0.3, 0.5, and 0.7, and the points are randomly sampled to 512, 256, and 128 points, respectively. Similar to~\cite{M2Track}, we enlarge the target box predicted in the previous frame by 2 meters to obtain the search area in the current frame. We utilize the targetness prediction operation~\cite{M2Track} as a pre-process in our tracking framework. Our tracking model is trained for 100 epochs using the Adam optimizer. The initial learning rate is set to 0.001 and is divided by 5 every 20 epochs. At the beginning of tracking, we use the target points lying inside the given box to initialize $\mathcal {T}$. Our project will be released at \textcolor{pink}{\textit{https://github.com/zjwhit/SCVTrack}}.

\vspace{2mm}\noindent
\textbf{Benchmarks and metrics.} 
We evaluate our algorithm on KITTI~\cite{KITTI}, NuScenes~\cite{NuScenes}, and Waymo Open Dataset (WOD)~\cite{Waymo}. KITTI consists of 21 training and 29 test sequences. We split the training set into train/validation/test splits as the test labels are inaccessible, following~\cite{SC3D, M2Track}. NuScenes comprises 1000 scenes, which are divided into train/validation/test sets. Following~\cite{BAT, M2Track}, we use the ``train\_track" split of the train set to train our model and test it on the validation set. WOD contains 1150 scenes, of which 798/202/150 scenes are used for training/validation/testing, respectively. We evaluate our method on WOD following two protocols: Protocol I~\cite{CXTrack}, where we directly test the KITTI pre-trained model on the validation set to evaluate generalization; Protocol II~\cite{M2Track}, in which the model is trained on the training set and evaluated on the validation set. 

We use success and precision as metrics to evaluate tracking performance. Specifically, we report the Area Under Curve (AUC) of the success plot and precision plot.

\subsection{Ablation Studies}
To analyze the effect of each component in SCVTrack, we conduct ablation experiments with six variants of SCVTrack: 1) the baseline (BL) model removing the shape completion mechanism and box refinement module from SCVTrack; 2) the variant using a naive shape completion mechanism without considering the point cloud quality into BL; 3) the variant performing quality-aware shape completion based on BL; 4) the variant that adopts the box refinement module based on the second variant; 5) our intact model; 6) the variant performing tracking with point features instead of voxelized features. This variant directly uses the attention-based method to process the point features and adopts an MLP head to regress the target box based on the output point features. Table~\ref{Tab:Ablation} presents the experimental results.

\newcommand{\tabincell}[2]{\begin{tabular}{@{}#1@{}}#2\end{tabular}}
\begin{table}[t!]
    \centering
    \rowcolors{2}{white}{gray!20}
    \setlength\tabcolsep{3pt}
    \renewcommand\arraystretch{1.2}
    \resizebox{1.0\linewidth}{!}{
    \begin{tabular}{l|cccc}
        \toprule
        Variants & Car & Cyclist & Van & Trailer\\
        \midrule
        1) BL  & 63.0 \textbar\ 78.6 & 72.5 \textbar\ 93.3 & 51.9 \textbar\ 68.1 & 56.6 \textbar\ 57.8\\
        2) BL+NSC & 65.2 \textbar\ 78.0 & 73.6 \textbar\ 93.5 & 54.9 \textbar\ 70.1 & 57.6 \textbar\ 58.4\\
        3) BL+QASC & 66.7 \textbar\ 79.2 & 75.1 \textbar\ 93.8 & 56.1 \textbar\ 71.9 & \underline{58.5} \textbar\ 59.5 \\
        4) BL+NSC+BR & \underline{67.0} \textbar\ \underline{79.6} & \underline{75.3} \textbar\ \underline{93.9} & \underline{57.8} \textbar\ \underline{72.1} &  58.4 \textbar\ \underline{59.6} \\
        5) BL+QASC+BR & \textbf{68.7} \textbar\ \textbf{81.9} & \textbf{77.4} \textbar\ \textbf{94.4} & \textbf{58.6} \textbar\ \textbf{72.8} &  \textbf{59.5} \textbar\ \textbf{60.1}\\
        6) Ours w/o Vox. & 64.5 \textbar\ \underline{79.6} & 74.5 \textbar\ 93.6 & 55.2 \textbar\ 71.0 & 57.5 \textbar\ 58.9\\
        \bottomrule
    \end{tabular}}
    \caption{Ablation study results on the car, cyclist, van, and trailer categories. BL refers to the baseline model. NSC denotes naive shape completion. QASC is quality-aware shape completion. BR refers to box refinement. Vox. means voxelization. The best and second-best scores are marked in bold and underline, respectively. Success~\textbar~Precision are reported.}
    \label{Tab:Ablation}
\end{table}
\vspace{2mm}\noindent
\textbf{Effect of the shape completion mechanism.}
The comparisons between the first three variants show that both the naive shape completion and quality-aware shape completion mechanisms can boost tracking performance. It manifests that performing shape completion in the raw point cloud space is an effective way to address the challenges of sparsity and incompleteness. 

\begin{table}[t]
    \centering
    \setlength\tabcolsep{3pt}
    \rowcolors{2}{white}{gray!20}
    \renewcommand\arraystretch{1.2}
    \resizebox{1.0\linewidth}{!}{
    \begin{tabular}{l|ccccc}
        \toprule
         & Car & Pedestrian & Van & Cyclist & Mean\\
        \midrule
        SC3D & 41.3 \textbar\ 57.9 & 18.2 \textbar\ 37.8& 40.4 \textbar\ 47.0 & 41.5 \textbar\ 70.4 & 31.2 \textbar\ 48.5\\
        P2B & 56.2 \textbar\ 72.8& 28.7 \textbar\ 49.6& 40.8 \textbar\ 48.4& 32.1 \textbar\ 44.7 & 42.4 \textbar\ 60.0\\
        LTTR & 65.0 \textbar\ 77.1& 33.2 \textbar\ 56.8& 35.8 \textbar\ 45.6& 66.2 \textbar\ 89.9 & 48.7 \textbar\ 65.8\\
        BAT & 60.5 \textbar\ 77.7 & 42.1 \textbar\ 70.1& 52.4 \textbar\ 67.0& 33.7 \textbar\ 45.4 & 51.2 \textbar\ 72.8\\
        PTT & 67.8 \textbar\ 81.8& 44.9 \textbar\ 72.0& 43.6 \textbar\ 52.5& 37.2 \textbar\ 47.3 & 55.1 \textbar\ 74.2\\
        PTTR & 65.2 \textbar\ 77.4& 50.9 \textbar\ 81.6& 52.5 \textbar\ 61.8& 65.1 \textbar\ 90.5 & 58.4 \textbar\ 77.8\\ 
        V2B & 70.5 \textbar\ 81.3 & 48.3 \textbar\ 73.5& 50.1 \textbar\ 58.0& 40.8 \textbar\ 49.7 & 58.4 \textbar\ 75.2\\
        CMT & 70.5 \textbar\ 81.9 & 49.1 \textbar\ 75.5 & 54.1 \textbar\ 64.1 & 55.1 \textbar\ 82.4 & 59.4 \textbar\ 77.6\\
        STNet & 72.1 \textbar\ 84.0 & 49.9 \textbar\ 77.2 & 58.0 \textbar\ 70.6 & 73.5 \textbar\ 93.7 & 61.3 \textbar\ 80.1\\
        M$^2$T & 65.5 \textbar\ 80.8 & 61.5 \textbar\ 88.2& 53.8 \textbar\ 70.7& 73.2 \textbar\ 93.5 & 62.9 \textbar\ 83.4\\
        TAT & 72.2 \textbar\ 83.3 & 57.4 \textbar\ 84.4 & 58.9 \textbar\ 69.2 & 74.2 \textbar\ 93.9 & 64.7 \textbar\ 82.8\\
        CXT & 69.1 \textbar\ 81.6 & 67.0 \textbar\ 91.5 & 60.0 \textbar\ 71.8 & 74.2 \textbar\ 94.3 & 67.5 \textbar\ 85.3 \\
        MTM & \underline{73.1} \textbar\ \underline{84.5} & \textbf{70.4} \textbar\ \textbf{95.1} & \underline{60.8} \textbar\ \textbf{74.2} & \underline{76.7} \textbar\ \textbf{94.6} & \textbf{70.9} \textbar\ \textbf{88.4}\\
        MBPT & \textbf{73.4} \textbar\ \textbf{84.8} & \underline{68.6} \textbar\ \underline{93.9} & \textbf{61.3} \textbar\ 72.7 & \underline{76.7} \textbar\ 94.3 & \underline{70.3} \textbar\ \underline{87.9}\\
        \textbf{Ours} & 68.7 \textbar\ 81.9 & 62.0 \textbar\ 89.1 & 58.6 \textbar\ \underline{72.8}& \textbf{77.4} \textbar\ \underline{94.4} & 65.1 \textbar\ 84.5\\
        \bottomrule
    \end{tabular}}
    \caption{Experimental results on KITTI. ``Mean” indicates the average results of the four categories.}
    \label{Tab:KITTI}
\end{table}

\begin{table}[t]
    \centering
    \rowcolors{2}{white}{gray!20}
    \setlength\tabcolsep{3pt}
    \renewcommand\arraystretch{1.2}
    \resizebox{1.0\linewidth}{!}{
    \begin{tabular}{l|ccccc}
        \toprule
         & \tabincell{c}{Car\\$\le \!\!150$} & \tabincell{c}{Pedestrian\\$\le \!\!100$} & \tabincell{c}{Van\\$\le \!\!150$} & \tabincell{c}{Cyclist\\$\le \!\!100$} & Mean\\
        \midrule
        SC3D & 37.9 \textbar\ 53.0 & 20.1 \textbar\ 42.0 & 36.2 \textbar\ 48.7 & 50.2 \textbar\ 69.2 & 32.7 \textbar\ 49.4\\
        P2B & 56.0 \textbar\ 70.6 & 33.1 \textbar\ 58.2 & 41.1 \textbar\ 46.3 & 24.1 \textbar\ 28.3 & 47.2 \textbar\ 63.5\\
        BAT & 60.7 \textbar\ 75.5 & 48.3 \textbar\ 77.1 & 41.5 \textbar\ 47.4 & 25.3 \textbar\ 30.5 & 54.3 \textbar\ 71.9\\
        V2B & \underline{64.7} \textbar\ \underline{77.4} & 50.8 \textbar\ 74.2 & 46.8 \textbar\ 55.1 & 30.4 \textbar\ 37.2 & 58.0 \textbar\ 73.2\\
        M$^2$T & 61.7 \textbar\ 75.9 & \underline{58.3} \textbar\ \underline{85.4} & \underline{50.2} \textbar\ \underline{68.5} & \underline{68.9} \textbar\ \underline{91.2} & \underline{59.3} \textbar\ \underline{77.6}\\ 
        \textbf{Ours} & \textbf{64.8} \textbar\ \textbf{77.7} & \textbf{60.1} \textbar\ \textbf{88.6} & \textbf{52.8} \textbar\ \textbf{70.5} & \textbf{70.2} \textbar\ \textbf{92.8} & \textbf{62.0} \textbar\ \textbf{80.1}\\
        \bottomrule
    \end{tabular}}
    \caption{Experimental results on sparse scenes of KIITI.}
    \label{Tab:Sparse scenes}
\end{table}
\noindent
\vspace{2mm}\textbf{Effect of the quality evaluator.}
The performance gaps between the second and third variants and between the fourth and fifth variants demonstrate that the quality evaluator can substantially improve the quality of the synthetic target representation and further improve tracking performance.

\vspace{2mm}\noindent
\textbf{Effect of the box refinement.}
Compared with the second and third variants, the fourth and fifth variants obtain large performance gains in the car and cyclist categories, respectively. It demonstrates the effectiveness of the box refinement guided by the synthetic target representation $\mathcal T$. 

\vspace{2mm}\noindent
\textbf{Effect of the voxelized tracking pipeline.}
Compared with our intact model, performing relation modeling and tracking with the imbalanced point features instead of the voxelized features, i.e., the sixth variant, results in substantial performance drops on these three categories. It validates that our voxelized tracking pipeline can deal with the aforementioned imbalance issue successfully but the point-feature-based tracking pipeline cannot.

\subsection{Quantitative Results}
We quantitatively compare our algorithm with state-of-the-art trackers on KITTI, NuScenes, and WOD. The trackers involved in the comparison include SC3D~\cite{SC3D}, P2B~\cite{P2B}, LTTR~\cite{LTTR}, PTT~\cite{PTT}, PTTR~\cite{PTTR} V2B~\cite{V2B}, BAT~\cite{BAT}, STNet~\cite{STNet}, M$^2$T~\cite{M2Track}, CMT~\cite{CMT}, TAT~\cite{TAT}, CXT~\cite{CXTrack}, MTM~\cite{MTM}, and MBPT~\cite{MBPT}. We discuss the results per dataset.

\begin{table*}[t]
    \centering
    \rowcolors{3}{gray!20}{white}
    \renewcommand\arraystretch{1.2}
    \resizebox{1.0\linewidth}{!}{
    \begin{tabular}{l|cccc|cccc|c}
        \toprule
         & \multicolumn{4}{c|}{Vehicle} & \multicolumn{4}{c|}{Pedestrian} & \multirow{2}{*}{Mean} \\
        & Easy & Medium & Hard & Mean & Easy & Medium & Hard & Mean & \\
        \midrule
        BAT$^{\dag}$ &  61.0 \textbar\ 68.3 & 53.3 \textbar\ 60.9 & 48.9 \textbar\ 57.8 & 54.7 \textbar\ 62.7 & 19.3 \textbar\ 32.6 & 17.8 \textbar\ 29.8 & 17.2 \textbar\ 28.3 & 18.2 \textbar\ 30.3 & 34.1 \textbar\ 44.4\\
        V2B$^{\dag}$ &  64.5 \textbar\ 71.5 & 55.1 \textbar\ 63.2 & 52.0 \textbar\ 62.0 & 57.6 \textbar\ 65.9 & 27.9 \textbar\ 43.9 & 22.5 \textbar\ 36.2 & 20.1 \textbar\ 33.1 & 23.7 \textbar\ 37.9 & 38.4 \textbar\ 50.1\\
        STNet$^{\dag}$ &  65.9 \textbar\ \underline{72.7} & \underline{57.5} \textbar\ \underline{66.0} & \underline{54.6} \textbar\ \underline{64.7} & \underline{59.7} \textbar\ \underline{68.0} & 29.2 \textbar\ 45.3 & 24.7 \textbar\ 38.2 & 22.2 \textbar\ 35.8 & 25.5 \textbar\ 39.9 & 40.4 \textbar\ 52.1\\
        TAT$^{\dag}$ &  \underline{66.0} \textbar\ 72.6 & 56.6 \textbar\ 64.2 & 52.9 \textbar\ 62.5 & 58.9 \textbar\ 66.7 & 32.1 \textbar\ 49.5 & 25.6 \textbar\ 40.3 & 21.8 \textbar\ 35.9 & 26.7 \textbar\ 42.2 & 40.7 \textbar\ 52.8 \\
        CXT$^{\dag}$ & 63.9 \textbar\ 71.1 & 54.2 \textbar\ 62.7 & 52.1 \textbar\ 63.7 & 57.1 \textbar\ 66.1 & \textbf{35.4} \textbar\ \underline{55.3} & \underline{29.7} \textbar\ \underline{47.9} & \underline{26.3} \textbar\ \underline{44.4} & \underline{30.7} \textbar\ \underline{49.4} & \underline{42.2} \textbar\ \underline{56.7}\\
        \textbf{Ours}$^{\dag}$ & \textbf{67.9} \textbar\ \textbf{76.1} & \textbf{58.8} \textbar\ \textbf{67.1} & \textbf{56.2} \textbar\ \textbf{65.3} & \textbf{61.3} \textbar\ \textbf{69.8} & \underline{35.3} \textbar\ \textbf{54.5} & \textbf{31.4} \textbar\ \textbf{48.8} & \textbf{29.4} \textbar\ \textbf{46.3} & \textbf{32.2} \textbar\ \textbf{50.0} & \textbf{44.8} \textbar\ \textbf{58.6}\\
        \bottomrule
    \end{tabular}}
    \caption{Experimental results of different methods on WOD following Protocol I. $\dag$ denotes the model is pre-trained on KITTI and directly evaluated on WOD validation split. These tracking results measure the generalization ability.}
    \label{tab:waymo}
\end{table*}

\begin{table*}[t]
    \centering
    \rowcolors{3}{white}{gray!20}
    \renewcommand\arraystretch{1.2}
    \begin{tabular}{>{\small}l|>{\small}c>{\small}c>{\small}c>{\small}c>{\small}c>{\small}c|>{\small}c>{\small}c>{\small}c}
        \toprule
         & \multicolumn{6}{c|}{NuScenes} & \multicolumn{3}{c}{Waymo Open Dataset}\\
        & Car & Pedestrian & Truck & Trailer & Bus & Mean & Vehicle & Pedestrian & Mean\\
        \midrule
        SC3D & 22.3 \textbar\ 21.9& 11.3 \textbar\ 12.7& 30.7 \textbar\ 27.7& 35.3 \textbar\ 28.1& 29.4 \textbar\ 24.1& 20.7 \textbar\ 20.2& -- & -- & --\\
        P2B & 38.8 \textbar\ 43.2& 28.4 \textbar\ 52.2& 43.0 \textbar\ 41.6& 49.0 \textbar\ 40.1& 33.0 \textbar\ 27.4& 36.5 \textbar\ 45.1& 28.3 \textbar\ 35.4& 15.6 \textbar\ 29.6& 24.2 \textbar\ 33.5\\
        BAT & 40.7 \textbar\ 43.3& 28.8 \textbar\ 53.3& 45.3 \textbar\ 42.6& 52.6 \textbar\ 44.9& 35.4 \textbar\ 28.0& 38.1 \textbar\ 45.7& 35.6 \textbar\ 44.2& 22.1 \textbar\ 36.8& 31.2 \textbar\ 41.8\\
        M$^2$T & \underline{55.9} \textbar\ \underline{65.1}& \underline{32.1} \textbar\ \underline{60.9}& \underline{57.4} \textbar\ \underline{59.5}& \underline{57.6} \textbar\ \underline{58.3}& \underline{51.4} \textbar\ \underline{51.4}& \underline{49.2} \textbar\ \underline{62.7}& \underline{43.6} \textbar\ \underline{61.6}& \underline{42.1} \textbar\ \underline{67.3}& \underline{43.1} \textbar\ \underline{63.5}\\
        \textbf{Ours} & \textbf{58.9} \textbar\ \textbf{67.7}& \textbf{34.5} \textbar\ \textbf{61.5}& \textbf{60.6} \textbar\ \textbf{61.4}& \textbf{59.5} \textbar\ \textbf{60.1}& \textbf{54.3} \textbar\ \textbf{53.6}& \textbf{52.1} \textbar\ \textbf{64.7}& \textbf{46.4} \textbar\ \textbf{63.0}& \textbf{44.1} \textbar\ \textbf{68.2}& \textbf{45.7} \textbar\ \textbf{64.7}\\
        \bottomrule
    \end{tabular}
    \caption{Experimental results of different methods on Nuscenes and WOD. These methods are trained on the training split of the Nuscenes or WOD benchmark and evaluated on the corresponding validation split.}
    \label{tab:nusenes}
\end{table*}

\begin{figure*}[t]
    \centering
    \includegraphics[width=1.0\textwidth]{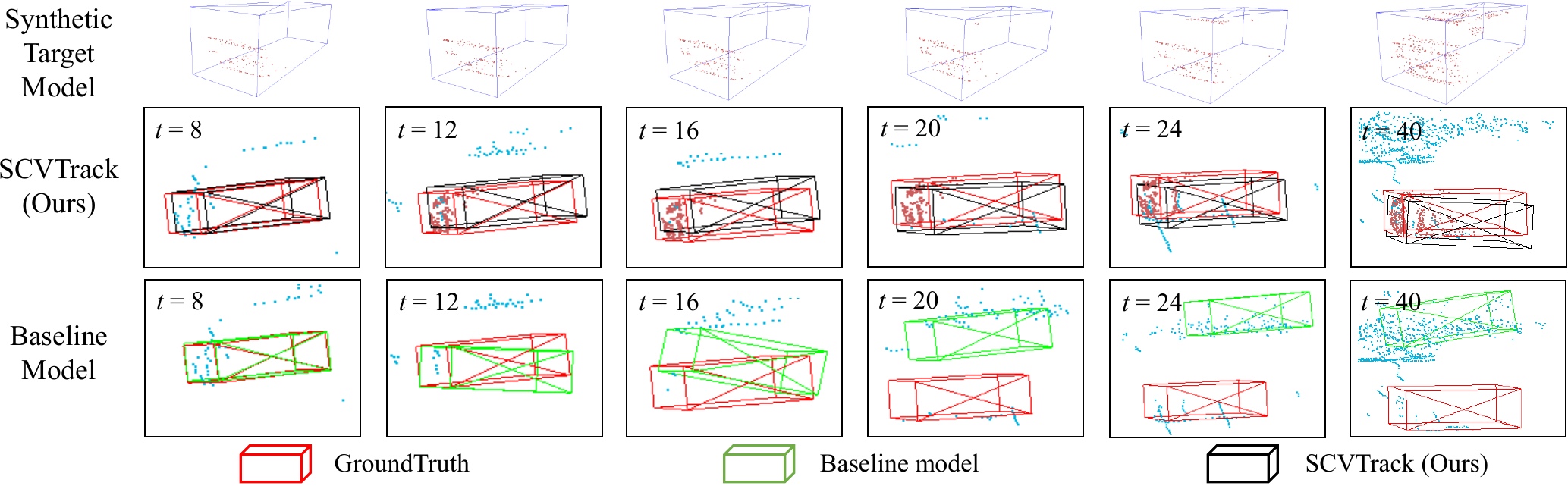}
    \caption{Qualitative comparisons between the variants w/ and w/o shape completion. Blue and red points refer to the raw points and fused points in each frame. We can observe that the shape completion mechanism helps SCVTrack successfully track the target in the extremely sparse scene, even though the synthetic target representation is not satisfactorily dense and complete.}
\label{Fig:visualization}
\end{figure*}

\begin{table}[t]
    \centering
    \renewcommand\arraystretch{1.2}
    \begin{tabular}{>{\small}c|>{\small}c>{\small}c>{\small}c}
        \toprule
         & Pre-process & Shape completion & Pointnet++ \\
        Time & 1.3 ms & 11.1 ms & 10.6 ms \\
        \midrule
         & Voxelization & Relation modeling & Box refinement\\
        Time & 1.1 ms & 5.6 ms & 1.7 ms\\
        \bottomrule
    \end{tabular}
    \caption{Inference time of each component of our model.}
    \label{tab:time}
\end{table}

\vspace{2mm}\noindent
\textbf{KITTI.} Table~\ref{Tab:KITTI} reports the experimental results on KITTI. V2B and TAT opt for dense geometric feature learning and sparse feature aggregation to address the sparsity and incompleteness challenges, respectively. Compared with them, our algorithm achieves better tracking performance in most categories. Our SCVTrack also outperforms M$^2$T in all categories, demonstrating its effectiveness. CXT and MBPT are two recently proposed trackers with sophisticated transformer blocks for relation modeling and target localization and perform better than our approach.

\vspace{2mm}\noindent
\textbf{WOD.} We first evaluate our SCVTrack on WOD following Protocol I to evaluate its generalization ability. Table~\ref{tab:waymo} reports the experimental results. Compared with TAT and CXT, our SCVTrack achieves performance gains of 2.6\% in mean success and 1.9\% in mean precision. This comparison shows that our method obtains stronger generalization ability. We also evaluate SCVTrack on WOD following Protocol II. As shown in Table~\ref{tab:nusenes}, our SCVTrack achieves the best performance in both vehicle and pedestrian categories.

\vspace{2mm}\noindent
\textbf{NuScenes.} Table~\ref{tab:nusenes} reports the experimental results on NuScenes. Our SCVTrack achieves the best success and precision score in the five categories. Compared with M$^2$T, our SCVTrack achieves performance gains of 2.9\% in mean success and 2.0\% in mean precision, demonstrating the effectiveness of our SCVTrack.

\vspace{2mm}\noindent
\textbf{Quantitative results in sparse scenes.} To investigate the effectiveness of our method in sparse scenes, we follow V2B to evaluate the performance in the sparse scenes (Car~$\le 150$, Pedestrian~$\le 100$, Van~$\le 150$, and Cyclist $\le 100$) of KITTI, as shown in Table~\ref{Tab:Sparse scenes}.
Our SCVTrack performs favorably against the other methods in all categories.

\vspace{2mm}\noindent
\textbf{Tracking speed.} We measure the average tracking speed on Car of KITTI on an Nvidia RTX3090 GPU, which is about 31 FPS and reaches \textbf{real-time} speed. The average inference time per frame is 31.4 ms. Table~\ref{tab:time} reports the detailed time consumption.

\subsection{Qualitative Results}
To further investigate the effectiveness of the shape completion mechanism, we visualize the tracking results of our SCVTrack and baseline model and the synthetic target representation in an extremely sparse scene, as shown in Figure~\ref{Fig:visualization}. Although the synthetic target representation is not satisfactorily dense due to the extremely sparse point clouds, our SCVTrack keeps tracking the target successfully. By contrast, the baseline model loses the target then the point clouds become extremely sparse and incomplete. 

\begin{figure}[t]
    \centering
    \includegraphics[width=1.0\columnwidth]{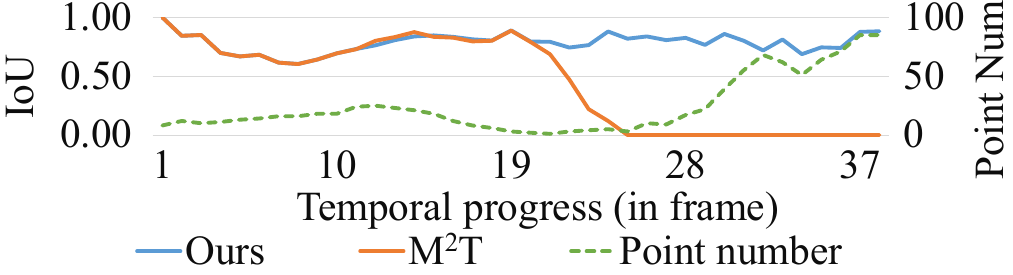}
    \caption{Results of M$^2$T and ours on a sparse scene.}
\label{Fig:Iou_plot}
\end{figure}

Figure~\ref{Fig:Iou_plot} compares the tracking results of our method and M$^2$T~\cite{M2Track} in a sparse scene. They can both track the target at the beginning. M$^2$T loses the target at about the $20^{th}$ frame (the target point cloud becomes quite sparse), while our method keeps tracking the target accurately.

\section{Conclusion}
In this work, we have presented a robust voxelized tracking framework with shape completion, named SCVTrack. Our SCVTrack constructs a dense and complete point cloud depicting the shape of the target precisely, i.e., the synthetic target representation, through shape completion and performs tracking with it in a voxelized manner. Specifically, we design a quality-aware shape completion mechanism, which can effectively alleviate the adverse effect of noisy historical predictions in shape completion. We also develop a voxelized relation modeling module and box refinement module to improve tracking performance. The proposed SCVTrack achieves favorable performance against state-of-the-art algorithms on three popular 3D tracking benchmarks.

\section*{Acknowledgements}
This work was supported in part by the National Natural Science Foundation of China (Grant NO. U2013210, 62006060, 62372133, 62176077), in part by the Guangdong Basic and Applied Basic Research Foundation under Grant (Grant NO. 2022A1515010306),  in part by Shenzhen Fundamental Research Program  (Grant NO. JCYJ20220818102415032), in part by the Shenzhen Key Technical Project (NO. 2022N001, 2020N046), in part by the Guangdong International Science and Technology Cooperation Project (NO. 20220505), in part by the Guangdong Provincial Key Laboratory of Novel Security Intelligence Technologies (NO. 2022B1212010005), and in part by the Key Research Project of Peng Cheng Laboratory (NO. PCL2023A08).

\appendix

\section{Additional Experiment Details}
\begin{figure}[!h]
    \centering
    \includegraphics[width=0.9\columnwidth]{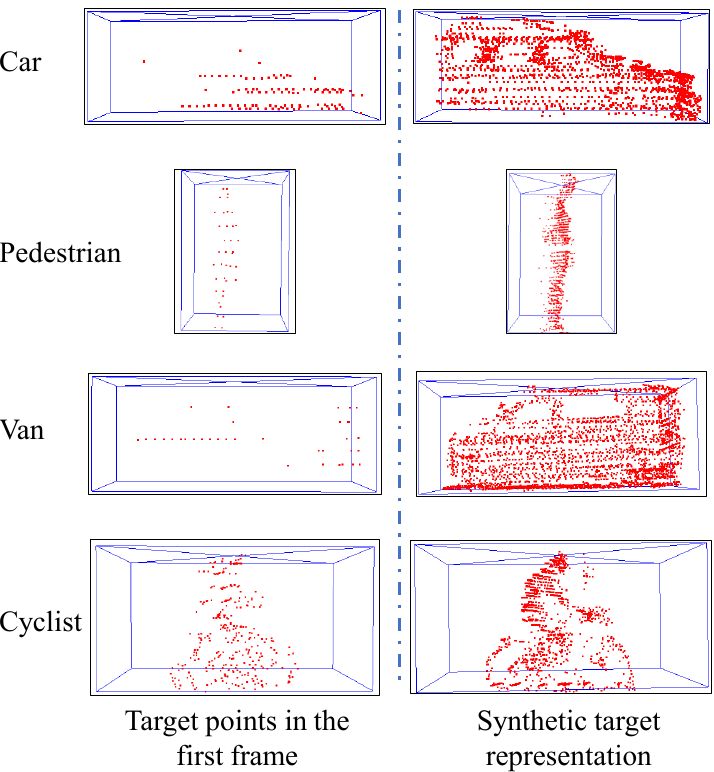}
    \caption{Visualization of the target points in the first frame and the synthetic target representation. We can observe that the former is sparse and incomplete while the latter is dense and complete. This visualization shows that the synthetic target representation can characterize the shape of the target object precisely.}
    \label{Fig:Synthetic_target_representation}
\end{figure}

\begin{figure*}[!h]
    \centering
    \includegraphics[width=0.965\textwidth]{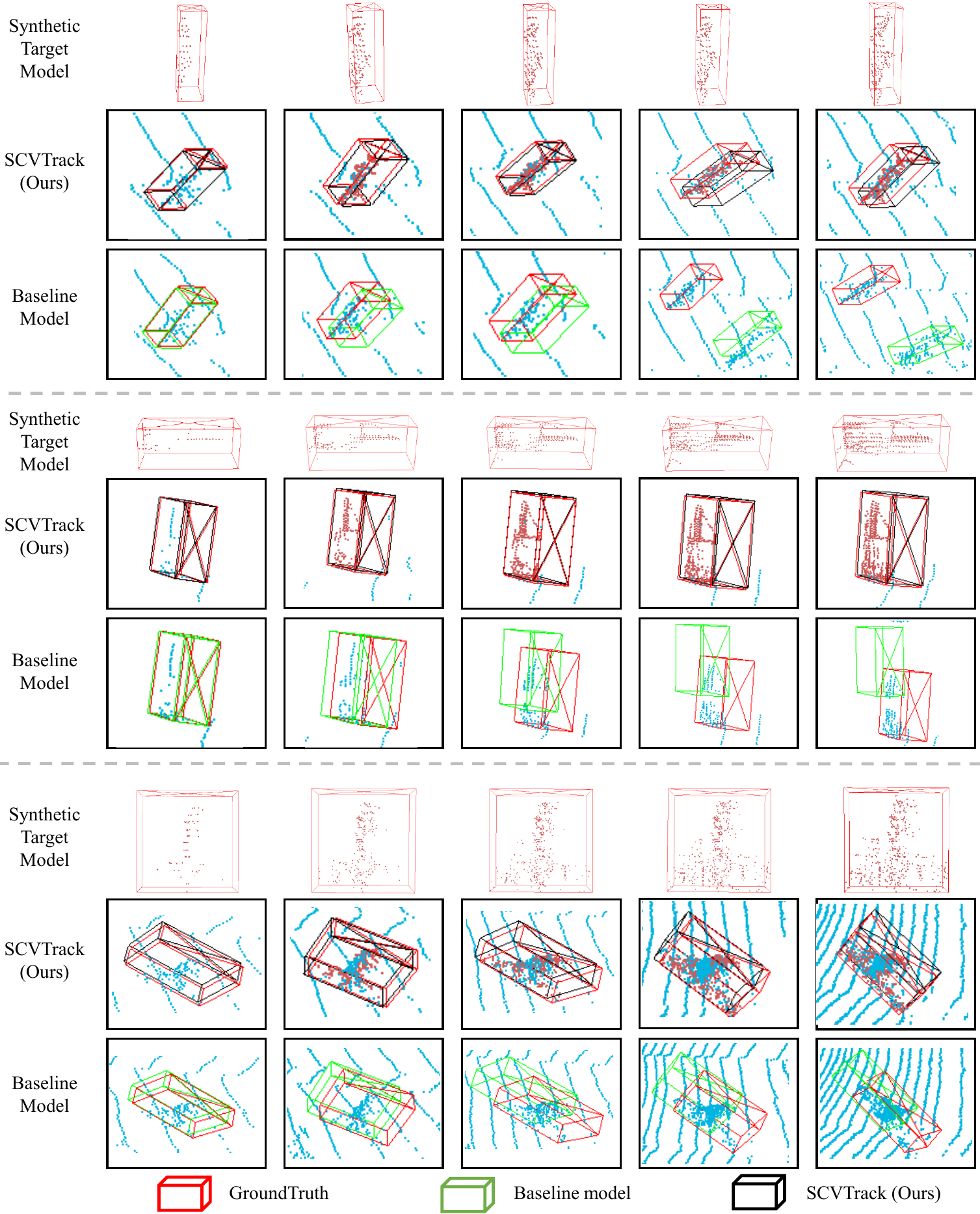}
    \caption{\textbf{Qualitative comparisons between the variants w/ and w/o shape completion.} Blue and red points refer to the raw points and fused points in each frame. The shape completion mechanism helps SCVTrack successfully track the target in these sparse scenes, demonstrating its effectiveness.}
    \label{Fig:Visual_comparison}
\end{figure*}
\subsubsection{Model details.}
As described in the main manuscript, we perform voxel-wise shape completion in our proposed quality-aware shape completion module, in which we divide the 3D space into voxel bins with a fixed size of 0.3 meters. During shape completion, we set an upper limit on the number of points within each voxel bin to prevent excessive imbalance in terms of point density across different areas within the completed point cloud. This upper limit is tuned to be 10 in our experiments. Unlike the voxelization manner for shape completion, for voxelized relation modeling, we uniformly partition the search area into a fixed number of voxel bins, which is $20\times 20\times 20$. This design ensures consistent computation complexity for processing the target object with different sizes. The hyperparameter $L$ is set to 2. Before predicting the coarse box $\mathcal {\hat B}_{t}$ with the MLP regression head, we first adopt four $1\times 1$ convolution layers (whose output dimensions are 256, 128, 32, and 4, respectively) to reduce the dimension of the enhanced feature $\mathcal {\tilde F}_{t}$. The following MLP regression head consisting of three linear layers (whose output dimensions are 256, 128, and 4, respectively) predicts $\Delta x, \Delta y, \Delta z, \Delta \theta$ w.r.t. $\mathcal {B}_{t-1}$ to estimate the coarse box $\mathcal {\hat B}_{t}$. The box refinement module consists of PointNet~\cite{PointNet} following a three-layer MLP (whose output dimensions are 512, 256, and 4, respectively), which predicts $\Delta x, \Delta y, \Delta z, \Delta \theta$ w.r.t. $\mathcal {\tilde B}_{t}$ to obtain the refined box $\mathcal B_{t}$.

\subsubsection{Training details.}
In our framework, the quality evaluator and the tracking model are end-to-end trained separately. We train the quality evaluator using the Adam optimizer~\cite{Adam} with a learning rate of 0.001 for 50 epochs. The tracking model is optimized with the Adam optimizer for 100 epochs. The initial learning rate is set to 0.001 and is divided by 5 every 20 epochs. Following M$^2$Track~\cite{M2Track}, we use data augmentations to improve the generalization of the tracking model, including the horizontal flipping of target points and bounding boxes, uniform rotations within -10° to 10° around their up-axes, and random translations using offsets from -0.3 to 0.3 meters. Our model is trained with four Nvidia RTX3090 GPUs.

\subsubsection{Evaluation Metrics.}
We employ the One Pass Evaluation (OPE)~\cite{wu2013online} method to evaluate the model. As mentioned in the main manuscript, we use success and precision to evaluate 3D tracking performance, which are calculated based on the overlap ratio (OR) and center location error (CLE) between 
the predicted and ground-truth boxes. The success plot measures the percentage of the frames in which OR is higher than a given threshold, and the precision plot measures the percentage of the frames whose CLE is lower than a given threshold. We report the Area Under the Curve (AUC) of the success plot (the OR threshold varying from 0 to 1) and the AUC of the precision plot (the CLE threshold varying from 0 to 2 meters) in our main manuscript.

\section{Additional Qualitative Results}
To obtain more insights into the pros and cons of our SCVTrack, we present more qualitative results in this section. We first visualize the synthetic target representation generated by our quality-aware shape completion module. As shown in Figure~\ref{Fig:Synthetic_target_representation}, compared with the target points in the first frame, the synthetic target representation is much more dense and complete, depicting the shape of the target object more precisely. To further investigate the effectiveness of our proposed shape completion mechanism, we visualize the tracking results of our SCVTrack across more categories on KITTI, as shown in Figure~\ref{Fig:Visual_comparison}. Blue and red points refer to the raw points and fused points in each frame. In these sparse scenes, it can be observed that our proposed shape completion mechanism leads to substantially better tracking results.

\bibliography{aaai24}
\end{document}